\pdfoutput=1

\documentclass[11pt]{article}

\usepackage[]{acl}

\usepackage{times}
\usepackage{latexsym}

\usepackage[T1]{fontenc}

\usepackage[utf8]{inputenc}

\usepackage{microtype}

\usepackage{graphicx}

\title{A Platform for Generating Educational Activities to Teach English as a Second Language}

\author{Aiala Rosá $\dagger$ \\
  \texttt{aialar@fing.edu.uy} \\
  \And
  Santiago Góngora $\dagger$ \\
  \texttt{sgongora@fing.edu.uy} \\
  \AND
  Juan Pablo Filevich  $\dagger$ \\
  \texttt{juan.filevich@fing.edu.uy} \\
  \And
  Ignacio Sastre $\dagger$\\
  \texttt{isastre@fing.edu.uy} \\
  \AND
  Laura Musto $\S$ \\
  \texttt{laura.musto@fic.edu.uy} \\
  \And
  Brian Carpenter $^*$ \\
  \texttt{bcarpent@iup.edu} \\
  \AND
  Luis Chiruzzo $\dagger$ \\
  \texttt{luischir@fing.edu.uy} \\
  \\
  $\dagger$ Facultad de Ingeniería, Universidad de la República, Uruguay \\
  $\S$ Facultad de Información y Comunicación, Universidad de la República, Uruguay \\
  $^*$ Indiana University of Pennsylvania, Indiana, PA, USA \\
}

\begin{document}
\maketitle
\begin{abstract}
We present a platform for the generation of educational activities oriented to teaching English as a foreign language. The different activities --games and language practice exercises-- are strongly based on Natural Language Processing techniques. 
The platform offers the possibility of playing out-of-the-box games, generated from resources created semi-automatically and then manually curated.
It can also generate games or exercises of greater complexity from texts entered by teachers, providing a stage of review and edition of the generated content before use. 
As a way of expanding the variety of activities in the platform, we are currently experimenting with image and text generation.
In order to integrate them and improve the performance of other neural tools already integrated, we are working on migrating the platform to a more powerful server.
In this paper we describe the development of our platform and its deployment for end users, discussing the challenges faced and how we overcame them, and also detail our future work plans.
\end{abstract}

\section{Introduction}

Although the application of language technologies to language teaching has been studied since the early days of Natural Language Processing~\cite{litman2016natural}, the development of educational platforms based on these approaches are not so common. 

One of the most studied NLP applications to language teaching has been the automatic correction of student work, mainly through some shared tasks as CoNLL~\cite{ng-etal-2014-conll} and BEA~\cite{bryant-etal-2019-bea}. 
These competitions led to the development of different corpora of human-corrected texts, which have been used for training correction systems.
Although there have been remarkable efforts on using NLP tools to generate educational activities (e.g. language practice exercises or games based on texts or words), the availability of platforms using them is scarce. 
When we talk about these platforms, we think of platforms built for and used by \textit{real users} (i.e. non-technical or academic users). 
An example
is Language Muse~\cite{burstein2013}, which allows to select a text from a catalog, or use own texts, and generate from them exercises for English that evaluate morphological, syntactic or semantic concepts.

NLP techniques allow the automatic or semi-automatic generation of exercises or games from texts selected by teachers, and also the generation of basic resources for language teaching, such as vocabularies, dictionaries, sentences and simple texts. 
In this paper we present an educational platform that integrates different activities for learning English\footnote{https://www.fing.edu.uy/inco/grupos/pln/cinacina/}, which we have carried out for over five years, within a collaboration with the National Public Education Administration of Uruguay. 
This collaboration began with the aim of supporting the teaching of English in rural schools, due to the lack of teachers trained in this language in schools in those areas. 
The development was carried out in constant interaction with its potential users, children and teachers, by an interdisciplinary team, integrated by experts in NLP and software development, and in Linguistics, in particular, language teaching. 
We also had permanent contributions from undergraduate students of Computer Science and students on Information and Communication, who developed prototypes of the different modules of the platform as course assignments, in addition to participating in the interactions with users. 
This modality of work, in collaboration with the community of users of the platform and with the contribution of undergraduate students, is presented in detail in~\citet{chiruzzo2022using}.

The platform can be used in two ways: On the one hand, it is possible to generate applications from pre-loaded resources, which are immediately available for use by children; on the other hand, it is possible to enter a text as input and generate activities based on it. This second modality includes a subsequent step of reviewing and editing the automatically generated activity, and is aimed at teachers with some proficiency in English.

Although the generated activities are persisted for future use, the platform does not save any personal data about the teachers that created them or the students that used them.
Additionally, the children do not need to register to use it, since the platform is free to use. 
We also plan to release the code soon.

In this paper we describe the general features of the platform and focus on the most recently integrated modules, that generate activities from an input text, as well as some modules under development, where we are experimenting with text and image generation. These modules use state-of-the-art techniques in the field. We also analyze the challenges faced while integrating NLP methods and resources into a platform for real users, and the difficulties to access appropriate servers with enough computing power.

\section{Related Work}

AI-tools for language learning have a positive effect on learners~\cite{van2021individual}, allowing a greater number of interactions around the language, which in turn can lead to positive perceptions of and a willingness to engage with the language in future learning~\cite{lin2021learning,li2022integration}. 
Teachers are using AI-tools to increase and support language learning, 
to bridge classroom and after class learning~\cite{yang2022current} and as a way to support interactions between the AI and learner~\cite{de2020effects}. 
This positive support then helps support students for future language learning~\cite{lin2021learning}.

The interest in educational applications has been present in the NLP area since its beginnings, and has been increasing in the Computational Linguistics community, leading to the creation in 2017 of the Association for Computational Linguistics Special Interest Group for building EDUcational applications (SIGEDU)\footnote{https://sig-edu.org/}. 
Within the area of Educational NLP, particular work has been done on the application of NLP to language teaching, a sub-area that has received the name Intelligent CALL or ICALL (CALL: Computer Assisted Language Learning)~\cite{ws-2014-nlp}. 
Our work, framed in this subarea of NLP for language teaching, focuses on the development of a platform of educational activities to support the teaching of English as a second language. 
There are few other developments of this type besides the aforementioned Language Muse~\cite{burstein2013} platform. A similar application but for multiple languages, REVITA, is presented in~\citet{katinskaia2018}. This tool generates a wide variety of exercises from texts and adapts the level to the user, based on previous answers.
The Lärka platform \cite{alfter2019larka}, for Swedish teaching, allows the generation of exercises from a corpus of real texts. It offers basic exercises of three types: parts of speech, syntactic relations and semantic roles.
Another platform for generating exercises from texts selected by teachers is presented in \citet{perez-cuadros-2017-multilingual}. This tool generates exercises to practice Spanish, Basque, English and French. The exercises can be fill-in-the-blanks, multiple choice and sentence reordering.
In~\citet{agirrezabal2019creating}, the development of activities for vocabulary learning from transformations of children's stories using NLP tools is described. A work that has an approach closely related to ours is~\citet{fenogenova2016automatic}, who build English exercises of different types, but mainly focused on learning collocations in English for more advanced students. 

Our platform has similarities with the aforementioned works, since it generates some classic language teaching exercises similar to those generated by them, but offers greater variety, as well as different didactic games. In addition, it combines text with images and sound.

\section{The Platform}

In the beginning our applications for English teaching were developed as isolated systems, each of them using the particular libraries, data sources and technologies needed.
As the number of applications grew, the difficulties to deploy them also grew.
One of the first iterations of this project was a desktop application which unified three games~\cite{FWW}: a crosswords generator, a word-search generator, and a battleship inspired game. We opted for a desktop application due to connectivity limitations in the rural schools we were interested in working with. 
Because of this decision, every time we wanted to test it with English students we had to install and configure it.
If we wanted to do an update (e.g. bug-fix or more content), we had to install the whole app again.
Although flawed, this app worked as a proof of concept that encouraged us to keep working in that direction, which lead to the platform we are presenting in this work.

\begin{figure}[h!]
    \centering
    \includegraphics[width=0.48\textwidth]{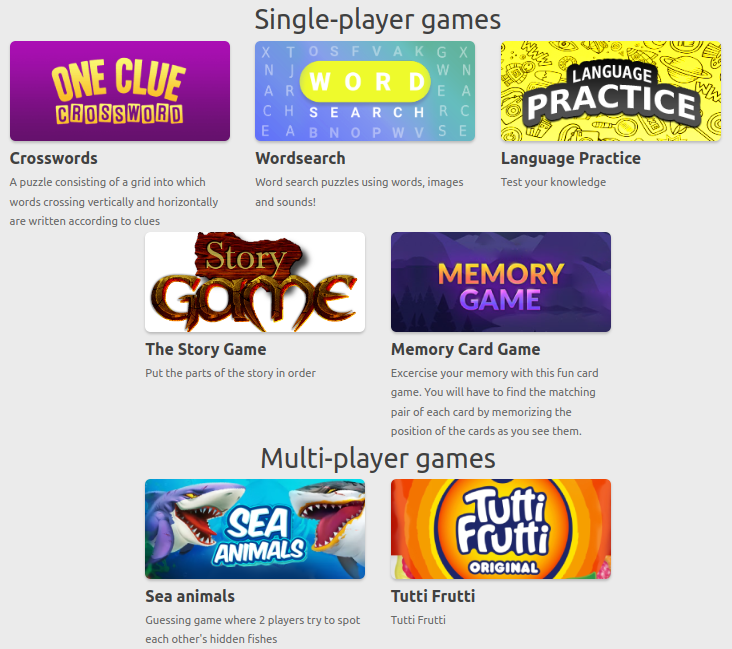}
    \caption{Screenshot of the landing screen of the platform.}
    \label{fig:cinacina}
\end{figure}

In 2021, we started to develop a unified web platform to integrate the different prototypes.
The tool integrates the different games and exercises, lets the users fix the errors that the NLP tools could introduce, and also serves as an environment to develop and deploy new tools and activities.
We abandoned the desktop application model and opted for a distributed approach, but trying to keep the bandwidth necessary to use the platform at a minimum.

In a similar way to digital distribution platforms 
like Steam\footnote{https://store.steampowered.com/about/}, our web platform acts as a gather point for the NLP applications for English teaching we are working on (see Fig.~\ref{fig:cinacina}).

These applications are pretty heterogeneous.
We developed some single-player games, as the aforementioned versions of well-known \textit{Word-search} and \textit{Crosswords}, the \textit{Memory} game, and the \textit{Story game}, where the student needs to reorder the sentences of a story.
We also developed two classic multiplayer games: \textit{Sea Animals}, based on the Battleship, and the \textit{Categories} word game. In addition, some basic multiple-choice exercises were also created for language practice. These activities can be generated by any user of the platform and are based on a list of words and definitions generated semi-automatically, with a subsequent manual curation process. Most of the above activities combine words and definitions with pictures, and in some cases also sound. The latter is essential for children to practice pronunciation, since the main audience of this platform are children that are learning with teachers who are not proficient in English.

The platform also allows to create more complex exercises, which work at sentence and text level. 
These activities are automatically generated from a text, which can be entered by the teacher. 
These activities include Question \& Answering exercises and new versions of Crossword, Word-search, and \textit{Story game}, this time automatically generated from an input text.
Given the complexity of these activities
it is possible to edit the generated material to correct errors or adapt the difficulty level. 

Having all the activities unified in a single platform allows us to exploit some shared characteristics, such as the data format used to communicate with the server or the vocabulary and definitions they rely on.

In what follows we will describe some aspects of these applications. First, a semi-automatically created dictionary of words and definitions is described, which is the basis for the generation of different activities. Since this resource was manually curated, this type of activities do not have a process of revision and editing by teachers. Even the children themselves can generate their own games or exercises and use them without prior review.
Secondly, we describe activities that are generated from texts entered by teachers. In this modality, teachers must log in to the platform, and they can review and edit the generated material before its use by the students.
Finally, some recent experiments on text and image generation are described.

\subsection{Resources Shared by Different Activities}

Many of the activities share some resources: lists of words classified in different categories, definitions associated with the words, an image bank. These resources were created semi-automatically, with a manual curation process that makes them usable to automatically generate activities that can be immediately used by children. These shared resources are described below.

\subsubsection{Words and Definitions}
\label{sec:vocabulary}

One of the main objectives of our platform is to dynamically generate games and exercises. 
Nowadays, following the recent NLP tendencies, the generation of content can be addressed through experimentation with neural networks and large language models (LLMs).
However these models have critical problems when used to build technologies for sensible populations like children, such as the prevalence of hallucinations or the possibility of generating unsuitable content~\cite{DBLP:journals/corr/abs-2202-03629,stochasticParrots}.
One way of filtering the output of the models and avoid the generation of unwanted language and content is using limited vocabularies.
Since the English-teaching program we are working with pays a lot of attention to basic words organized in categories, the usage of vocabularies seemed a suitable path, so a lot of effort to build appropriate vocabularies was done. 

We started with a word list for different learning levels, and we expanded it by applying some heuristics based on \textit{word embeddings} and term frequencies~\cite{FWW}. We used the average word embeddings of the words in each category (e.g. ``animals'' or ``food'') to obtain related words. Then we used the term frequencies obtained from Simple English Wikipedia\footnote{https://simple.wikipedia.org/wiki/Main\_Page} to filter words with a similar difficulty level to those included in the initial vocabulary. This process was followed by manual curation.
In order to get definitions for those words, we followed two approaches: scraping online dictionaries and extracting definition-definiendum pairs from Simple Wikipedia articles.  
Both for obtaining new words for each category and for obtaining definitions we used the word embedding of the category name (e.g. ``animal'' or ``food'') to mitigate ambiguity problems.
After performing another manual curation, we finally obtained the first working vocabulary, which was used in the initial version of the platform. Almost every activity available on the platform use the vocabulary in some way, trusting in its correctness and safeness, and making use of the carefully chosen tags.
These tags work as the previous mentioned categories, but in a less-rigid way: each word can have any number of tags.
This would allow the platform to use a single word in different contexts, like for a \textit{Sports quiz game} or a \textit{language practice} activity focused on ``jobs''.

The platform was then tested in classrooms with students and teachers.
The feedback obtained from this initial testing phase was used by English teachers to perform an in-depth curation of the vocabularies, improving the available words and their definitions.

\subsubsection{Using both Text and Images}

\begin{figure}[t!]
  \centering
  \includegraphics[width=0.5\textwidth]{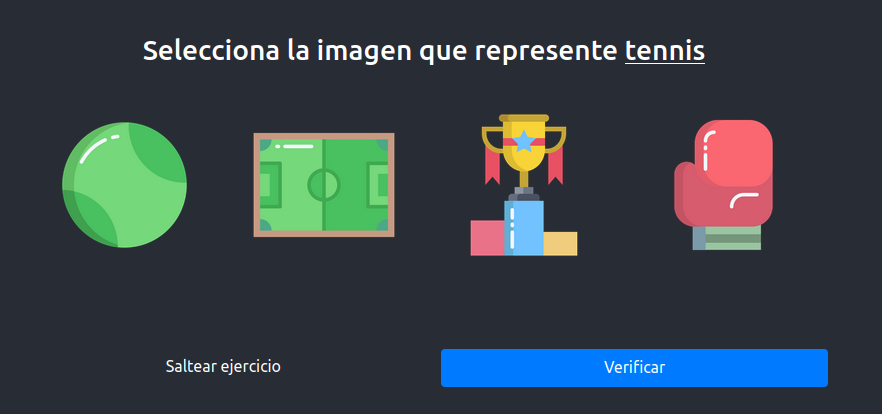}
  \caption{A game consisting of choosing the image related to the word ``tennis''.}
  \label{fig:image-text-exercise}
\end{figure}

When we tested the platform with teachers and students we received frequent requests about incorporating activities with images in the application, as they consider them to be highly engaging.
We created a database of images, mixing different open licensed image collections, which allows the creation of language exercises with a visual component, like the one shown in Fig.~\ref{fig:image-text-exercise}.

\subsection{Generating Activities from Text}
\label{sec:activitiesFromText}

The vocabulary is used when we want to generate out-of-the-box activities that use the previously curated words and definitions.
However, teachers may want to work with some particular text, like an article or story they are working with during class.
Because of this, we explored the generation of activities from an input text, adding an additional mode to the \textit{Crosswords}, \textit{Word-search}, and \textit{Story game} activities. We also included a new activity for creating \textit{Question \& Answering} exercises based on an input text. When activities are generated from an input text, the teacher must log in to the platform and can edit the generated activity before using it with their students.

The problem for the \textit{Crosswords} games is, in essence, the problem of extracting definitions from a text. 
However, since our platform is oriented to English learners, there is an additional requirement: the extracted definitions must be concise and simple.
The process of generating <word, clue> pairs from a text is divided into two stages: in a first stage a large set of manual patterns is applied to generate a large number of candidate pairs, and in a second stage a classifier is applied to determine whether each generated pair is a good clue to be included in a crossword puzzle.
For the second stage a manually annotated corpus was created, with around 2,200 examples, which was used to train different classifiers.
The best classifier found uses a DistilBERT model and has 0.78 F1-measure.
The crossword is completed with some words from the definition database described above.
This same definition extraction module is used to obtain words for a \textit{Word-search} game based on input text.

\begin{figure}[t!]
    \centering
    \includegraphics[width=0.48\textwidth]{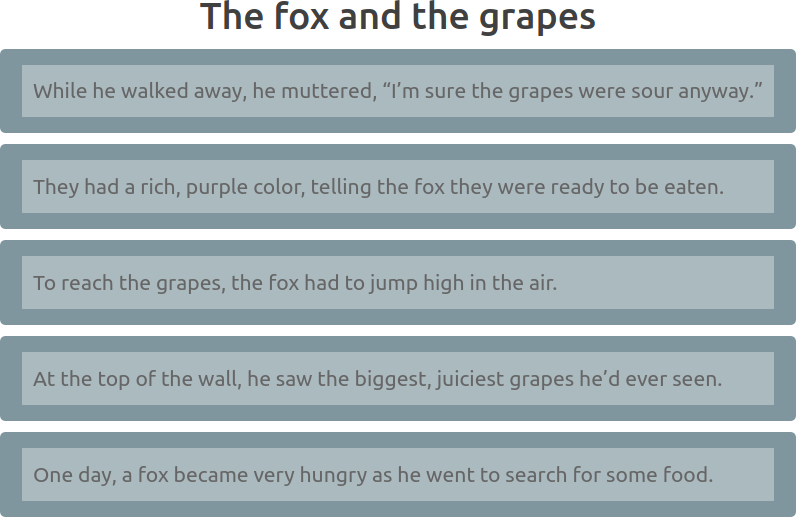}
    \caption{Story game based on the story ``The fox and the grapes'' with sentences shuffled, the student must put them in the correct order.}
    \label{fig:story_sample}
\end{figure}

In the case of the \textit{Story game}, the challenge is to find the most salient sentences from the input text.
This sentence selection is done by performing a ranking through token scoring~\cite{chiruzzo2022using}. 
The top-scored sentences are then shuffled and the player is asked to put them in order.
Fig.~\ref{fig:story_sample} shows an example of a shuffled story based on the classic short story ``The fox and the grapes''.

Finally, we created a module for the generation of Question \& Answering (Q\&A) exercises, which always requires an input text.
We worked on two different approaches for creating Q\&A exercises: a rule based approach, that uses semantic roles as the basic information to generate questions and answers~\cite{moron2021tool}, and a second approach based on a language model fine tuned with two large Q\&A corpora, SQUAD~\cite{rajpurkar2018know} and NewsQA~\cite{trischler2016newsqa}. 
The Q\&A module of the platform is based on the second approach~\cite{berger2022generation}. This module performs a four-stage processing: a pre-processing stage that applies a co-reference resolution tool; an answer candidate selection stage based on semantic role labeling; a question generation stage based on a transformer language model; and a post-processing stage that adjusts the format of the generated questions. 

\subsection{Image and Text Generation}

As we detailed in section \ref{sec:activitiesFromText}, one of our objectives is to have applications that generate activities from any text.
However, it is really hard to find images from our database that describe every possible text entered by the users (e.g. how can we illustrate, using a single image from our database, an animal doing a a particular action?).
In order to tackle this problem, we made some proofs of concept of the integration of text-to-image models, such as Stable Diffusion \cite{rombach2021highresolution}, and large language models, such as GPT-2 \cite{radford2019language}, to the platform.
These models are a good starting point for experimenting as they are both open source projects. 
However, they are also challenging to integrate in a production environment due to their high computing requirements.

\begin{figure}
  \centering
  \includegraphics[width=0.5\textwidth]{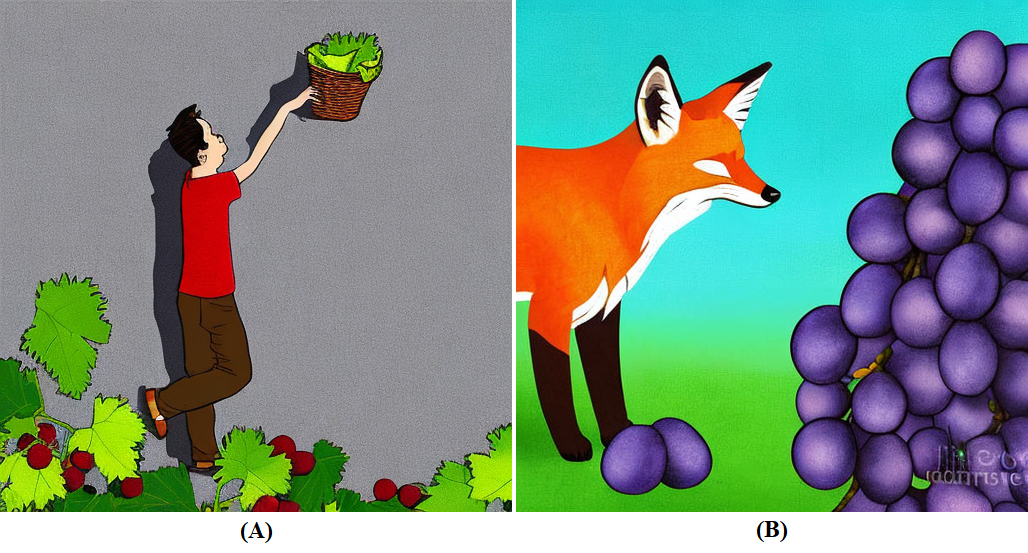}
  \caption{(A) Resulting image for \textit{"At the top of the wall, he saw the biggest, juiciest grapes he'd ever seen"}. (B) Resulting image with the co-reference resolution, changing the pronoun \textit{he} for \textit{a fox}.}
  \label{fig:corefereecomparison}
\end{figure}

We started by adding image generation to the \textit{Story game}. 
As mentioned before, this game consists of a list of shuffled sentences that are part of a story, and the objective is to order them chronologically. 
A new image is generated for each sentence, using the sentence as prompt followed by a description of the intended style. 
This method, however, was not working as expected because of the use of pronouns in the sentences, which hides part of the context needed for getting a coherent image. 
To solve this problem, we applied co-reference resolution using the sPacy library Coreferee\footnote{https://spacy.io/universe/project/coreferee}. 
The difference in the generated images can be appreciated in Figure~\ref{fig:corefereecomparison}.

\begin{figure}
  \centering
  \includegraphics[width=0.5\textwidth]{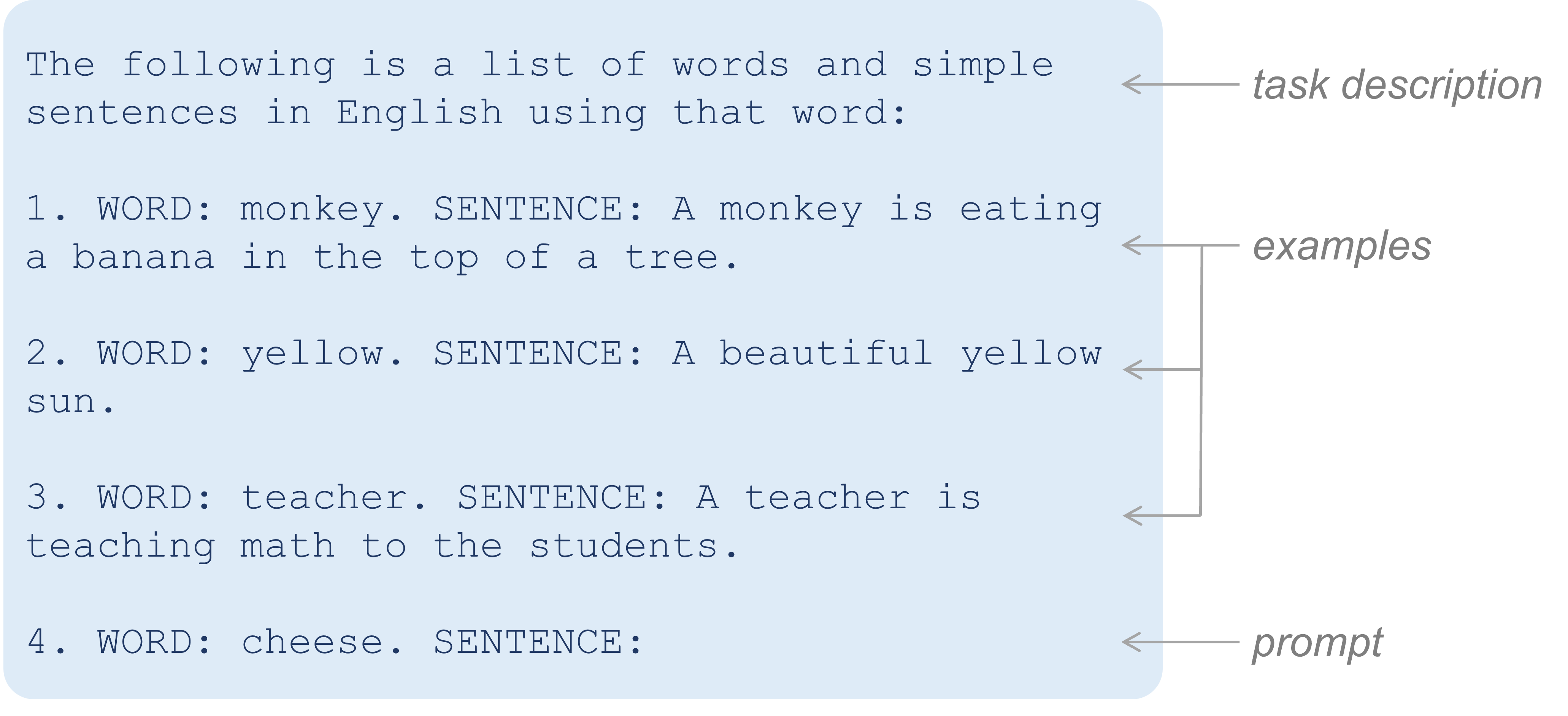}
  \caption{Prompt example for the generation of a sentence using GPT-2.}
  \label{fig:promptexample}
\end{figure}

Another proof of concept we developed was a multiple choice type of game, generating five English sentences and one image from one of these sentences. 
The objective of this game is to choose the sentence that the image is representing. 
We used a prompt-based few-shot learning approach \cite{llmsfewshot} for sentence generation.
The task is to create a short English sentence that contains a certain word. 
Three examples of $<$\textit{word, short sentence containing the word}$>$ pairs are provided, and the LLM has to complete the fourth pair given a random word from a fixed list of our vocabulary. 
An example of this approach is presented in Fig.~\ref{fig:promptexample}, where a prompt for the generation of a sentence containing the word \textit{cheese} is shown. 
Five sentences are generated in this way.
Finally, one is randomly selected to generate a representative image with a given style.

These experiments based on automatic generation of images and sentences are not yet integrated into the platform. One of the obstacles we face in order to integrate them is their high computing requirements. It should be noted that these would be activities that would always have a manual post-editing stage, in order to avoid generating material that would not be appropriate for children.

\section{Challenges and Limitations}

The development of applications for real users faces challenges that are usually not taken into account when focusing on particular NLP tasks and methods, generally oriented to improve results on specific datasets.
In this section we try to report our experience on these challenges.

\subsection{Challenges faced in the development and deployment of the platform} 
 
During the development and deployment of the platform we faced difficulties in integrating state-of-the-art methods into user-oriented applications, which ideally should provide answers in reasonable times, being used with commodity computers. Large language models, which today are the state of the art for many of the problems studied in NLP, are not always suitable for use under the aforementioned conditions. Once trained, these models require significant computing capacity and memory, so if these resources are not available, it becomes unfeasible to use them. 

The current version of our platform is functional on a server that does not have enough computing capacity or memory, nor does it have a GPU, to offer a reasonable usability of the activities that rely on language models, such as the Question \& Answering module, or the activities we plan to incorporate, based on text and image generation.

We are currently migrating the platform to a server which will allow better response times.
Some preliminary tests showed that we can lower the response times with respect to the previous server. In some tests response times drop from over one minute to a few seconds, which is a considerable improvement, even though this server, like the previous one, does not have a GPU integrated. 

\subsection{Using neural models in teaching environments}

Large Language Models, like GPT \cite{llmsfewshot}, ChatGPT,
LLaMA \cite{touvron2023llama} or FLAN \cite{wei2022finetuned} are showing great results on quantitative experiments (i.e. classic test sets or benchmarks). 
Although benchmarks seem to not being suitable for every AI problem \cite{raji2021ai,CC_SOTA}, LLMs go beyond and also generate the user the illusion of having the ability to understand language. However, whether or not they really can \textit{understand} human language is currently a hot object of debate \cite{bender-koller-2020-climbing,bowman2023things}.

Although the \textit{understanding} capabilities of LLMs can be a good addition to the applications in the platform (e.g. to check semantically correct answers instead of just performing pattern matching), what can be game-changing for us are their generative capabilities. 
With them, a big range of ideas are starting to become reachable as educational tools, like generating simple question-answering exercises from any text or complex activities like those involving Role-playing \cite{prager2019exploring}.
However, since these LLMs are trained on massive amounts of text with almost no revision, they are showing big problems like hallucination \cite{DBLP:journals/corr/abs-2202-03629}, spreading misinformation or generation of harmful comments \cite{stochasticParrots}.

As we mentioned in section \ref{sec:vocabulary}, these problems can be critical when working on applications for education, especially when children are involved.
That is why we think that every educational platform, like ours, should give the teachers the possibility to examine the generated content first, instead of just making the students interact freely with LLMs.
In other words, until the generated contents of LLMs are fully controllable and reliable, we should put the teachers in the loop in order to curate the activities generated for their students.
This makes harder the development of activities which require a quick answer by the system, like a chatbot or other types of interactive games.

\section{Conclusions}

In this paper we present a platform for teaching English, developed over five years, in collaboration with the National Public Education Administration of Uruguay. The platform offers, on the one hand, the possibility of generating games and exercises dynamically, consulting a database of semi-automatically generated and subsequently manually curated content, such as lists of words and definitions, and an image bank. Games include Word-Search and Crossword puzzles, as well as multi-player games, such as the Battleship and the Categories game. On the other hand, the platform includes modules for the generation of games and exercises from texts entered by teachers, offering them the possibility of reviewing and editing the generated activities. In this modality, it is possible to generate Question \& Answering exercises, Crossword puzzles associated to texts and the Story game, to order parts of a story.

Currently, we are experimenting with tools for the automatic generation of images, in order to illustrate some activities, such as the Story game, and also with the generation of sentences, to increase the variety of activities. The use of these tools is done with the necessary care to offer appropriate material for children.

Finally, we comment on some difficulties faced in developing a tool based on state-of-the-art NLP techniques, which require significant processing resources and human supervision of the generated content.

We plan in the future to have the platform installed on a more powerful server than the current one, and to integrate the activities that are in the experimental phase.

\bibliography{anthology,custom_full}

\begin{thebibliography}{33}
\expandafter\ifx\csname natexlab\endcsname\relax\def\natexlab#1{#1}\fi

\bibitem[{Agirrezabal et~al.(2019)Agirrezabal, Altuna, Gil~Vallejo, Goikoetxea,
  and Gonzalez~Dios}]{agirrezabal2019creating}
Manex Agirrezabal, Bego{\~n}a Altuna, Lara Gil~Vallejo, Josu Goikoetxea, and
  Itziar Gonzalez~Dios. 2019.
\newblock Creating vocabulary exercises through nlp.
\newblock \emph{Digital Humanities in the Nordic Countries. Proceedings, 2019}.

\bibitem[{Alfter et~al.(2019)Alfter, Borin, Pilán, Lindström~Tiedemann, and
  Volodina}]{alfter2019larka}
David Alfter, Lars Borin, Ildikó Pilán, Therese Lindström~Tiedemann, and
  Elena Volodina. 2019.
\newblock Lärka: From language learning platform to infrastructure for
  research on language learning.
\newblock In \emph{Selected papers from the CLARIN Annual Conference 2018,
  Pisa, 8-10 October 2018}, 159, pages 1--14. Linköping University Electronic
  Press.

\bibitem[{Bender et~al.(2021)Bender, Gebru, McMillan-Major, and
  Shmitchell}]{stochasticParrots}
Emily~M. Bender, Timnit Gebru, Angelina McMillan-Major, and Shmargaret
  Shmitchell. 2021.
\newblock \href {https://doi.org/10.1145/3442188.3445922} {On the dangers of
  stochastic parrots: Can language models be too big?}
\newblock In \emph{Proceedings of the 2021 ACM Conference on Fairness,
  Accountability, and Transparency}, FAccT '21, page 610–623, New York, NY,
  USA. Association for Computing Machinery.

\bibitem[{Bender and Koller(2020)}]{bender-koller-2020-climbing}
Emily~M. Bender and Alexander Koller. 2020.
\newblock \href {https://doi.org/10.18653/v1/2020.acl-main.463} {Climbing
  towards {NLU}: {On} meaning, form, and understanding in the age of data}.
\newblock In \emph{Proceedings of the 58th Annual Meeting of the Association
  for Computational Linguistics}, pages 5185--5198, Online. Association for
  Computational Linguistics.

\bibitem[{Berger et~al.(2022)Berger, Rischewski, Chiruzzo, and
  Ros{\'a}}]{berger2022generation}
Gonzalo Berger, Tatiana Rischewski, Luis Chiruzzo, and Aiala Ros{\'a}. 2022.
\newblock Generation of english question answer exercises from texts using
  transformers based models.
\newblock In \emph{2022 IEEE Latin American Conference on Computational
  Intelligence (LA-CCI)}. IEEE.

\bibitem[{Bowman(2023)}]{bowman2023things}
Samuel~R. Bowman. 2023.
\newblock \href {http://arxiv.org/abs/2304.00612} {Eight things to know about
  large language models}.

\bibitem[{Brown et~al.(2020)Brown, Mann, Ryder, Subbiah, Kaplan, Dhariwal,
  Neelakantan, Shyam, Sastry, Askell, Agarwal, Herbert-Voss, Krueger, Henighan,
  Child, Ramesh, Ziegler, Wu, Winter, Hesse, Chen, Sigler, Litwin, Gray, Chess,
  Clark, Berner, McCandlish, Radford, Sutskever, and Amodei}]{llmsfewshot}
Tom~B. Brown, Benjamin Mann, Nick Ryder, Melanie Subbiah, Jared Kaplan,
  Prafulla Dhariwal, Arvind Neelakantan, Pranav Shyam, Girish Sastry, Amanda
  Askell, Sandhini Agarwal, Ariel Herbert-Voss, Gretchen Krueger, Tom Henighan,
  Rewon Child, Aditya Ramesh, Daniel~M. Ziegler, Jeffrey Wu, Clemens Winter,
  Christopher Hesse, Mark Chen, Eric Sigler, Mateusz Litwin, Scott Gray,
  Benjamin Chess, Jack Clark, Christopher Berner, Sam McCandlish, Alec Radford,
  Ilya Sutskever, and Dario Amodei. 2020.
\newblock \href {https://doi.org/10.48550/ARXIV.2005.14165} {Language models
  are few-shot learners}.

\bibitem[{Bryant et~al.(2019)Bryant, Felice, Andersen, and
  Briscoe}]{bryant-etal-2019-bea}
Christopher Bryant, Mariano Felice, {\O}istein~E. Andersen, and Ted Briscoe.
  2019.
\newblock \href {https://doi.org/10.18653/v1/W19-4406} {The {BEA}-2019 shared
  task on grammatical error correction}.
\newblock In \emph{Proceedings of the Fourteenth Workshop on Innovative Use of
  NLP for Building Educational Applications}, pages 52--75, Florence, Italy.
  Association for Computational Linguistics.

\bibitem[{Burstein et~al.(2013)Burstein, Sabatini, Shore, Moulder, and
  Lentini}]{burstein2013}
J.~Burstein, J.~Sabatini, J.~Shore, B.~Moulder, and J.~Lentini. 2013.
\newblock A user study: Technology to increase teachers’ linguistic awareness
  to improve instructional language support for english language learners.
\newblock In \emph{Proceedings of the 2nd Workshop on Natural Language
  Processing for Improving Textual Accessibility (NLP4ITA)}, Atlanta, Georgia.
  Association for Computational Linguistics.

\bibitem[{Chiruzzo et~al.(2022)Chiruzzo, Musto, G{\'o}ngora, Carpenter,
  Filevich, and Ros{\'a}}]{chiruzzo2022using}
Luis Chiruzzo, Laura Musto, Santiago G{\'o}ngora, Brian Carpenter, Juan
  Filevich, and Aiala Ros{\'a}. 2022.
\newblock Using nlp to support english teaching in rural schools.
\newblock In \emph{Proceedings of the Second Workshop on NLP for Positive
  Impact (NLP4PI)}, pages 113--121.

\bibitem[{De~Haas et~al.(2020)De~Haas, Vogt, and Krahmer}]{de2020effects}
Mirjam De~Haas, Paul Vogt, and Emiel Krahmer. 2020.
\newblock The effects of feedback on children's engagement and learning
  outcomes in robot-assisted second language learning.
\newblock \emph{Frontiers in Robotics and AI}, 7:101.

\bibitem[{Fenogenova and Kuzmenko(2016)}]{fenogenova2016automatic}
Alena Fenogenova and Elizaveta Kuzmenko. 2016.
\newblock Automatic generation of lexical exercises.
\newblock In \emph{Proceedings of the Workshop on Computational Linguistics and
  Language Science}.

\bibitem[{Ji et~al.(2022)Ji, Lee, Frieske, Yu, Su, Xu, Ishii, Bang, Madotto,
  and Fung}]{DBLP:journals/corr/abs-2202-03629}
Ziwei Ji, Nayeon Lee, Rita Frieske, Tiezheng Yu, Dan Su, Yan Xu, Etsuko Ishii,
  Yejin Bang, Andrea Madotto, and Pascale Fung. 2022.
\newblock \href {http://arxiv.org/abs/2202.03629} {Survey of hallucination in
  natural language generation}.
\newblock \emph{CoRR}, abs/2202.03629.

\bibitem[{Jordanous(2022)}]{CC_SOTA}
Anna Jordanous. 2022.
\newblock \href
  {http://computationalcreativity.net/iccc22/papers/ICCC-2022\_paper\_134.pdf}
  {Should we pursue {SOTA} in computational creativity?}
\newblock In \emph{Proceedings of the 13th International Conference on
  Computational Creativity, Bozen-Bolzano, Italy, June 27 - July 1, 2022},
  pages 159--163. Association for Computational Creativity {(ACC)}.

\bibitem[{Katinskaia et~al.(2018)Katinskaia, Nouri, and
  Yangarber}]{katinskaia2018}
Anisia Katinskaia, Javad Nouri, and Roman Yangarber. 2018.
\newblock Revita: a language-learning platform at the intersection of its and
  call.
\newblock In \emph{Proceedings of the Eleventh International Conference on
  Language Resources and Evaluation (LREC 2018)}, Miyazaki, Japón. European
  Language Resources Association (ELRA).

\bibitem[{Li and Peng(2022)}]{li2022integration}
Bing Li and Miaomiao Peng. 2022.
\newblock Integration of an ai-based platform and flipped classroom
  instructional model.
\newblock \emph{Scientific Programming}, 2022.

\bibitem[{Lin and Mubarok(2021)}]{lin2021learning}
Chi-Jen Lin and Husni Mubarok. 2021.
\newblock Learning analytics for investigating the mind map-guided ai chatbot
  approach in an efl flipped speaking classroom.
\newblock \emph{Educational Technology \& Society}, 24(4):16--35.

\bibitem[{Litman(2016)}]{litman2016natural}
Diane Litman. 2016.
\newblock Natural language processing for enhancing teaching and learning.
\newblock In \emph{Thirtieth AAAI conference on artificial intelligence}.

\bibitem[{Mor{\'o}n et~al.(2021)Mor{\'o}n, Scocozza, Chiruzzo, and
  Ros{\'a}}]{moron2021tool}
Mart{\'\i}n Mor{\'o}n, Joaqu{\'\i}n Scocozza, Luis Chiruzzo, and Aiala
  Ros{\'a}. 2021.
\newblock A tool for automatic question generation for teaching english to
  beginner students.
\newblock In \emph{2021 40th International Conference of the Chilean Computer
  Science Society (SCCC)}, pages 1--5. IEEE.

\bibitem[{Ng et~al.(2014)Ng, Wu, Briscoe, Hadiwinoto, Susanto, and
  Bryant}]{ng-etal-2014-conll}
Hwee~Tou Ng, Siew~Mei Wu, Ted Briscoe, Christian Hadiwinoto, Raymond~Hendy
  Susanto, and Christopher Bryant. 2014.
\newblock \href {https://doi.org/10.3115/v1/W14-1701} {The {C}o{NLL}-2014
  shared task on grammatical error correction}.
\newblock In \emph{Proceedings of the Eighteenth Conference on Computational
  Natural Language Learning: Shared Task}, pages 1--14, Baltimore, Maryland.
  Association for Computational Linguistics.

\bibitem[{Percovich et~al.(2019)Percovich, Tosi, Chiruzzo, and Rosá}]{FWW}
Analía Percovich, Alejandro Tosi, Luis Chiruzzo, and Aiala Rosá. 2019.
\newblock \href {https://doi.org/10.1109/SCCC49216.2019.8966429} {Ludic
  applications for language teaching support using natural language
  processing}.
\newblock In \emph{2019 38th International Conference of the Chilean Computer
  Science Society (SCCC)}, pages 1--7.

\bibitem[{Perez and Cuadros(2017)}]{perez-cuadros-2017-multilingual}
Naiara Perez and Montse Cuadros. 2017.
\newblock \href {https://aclanthology.org/E17-3013} {Multilingual {CALL}
  framework for automatic language exercise generation from free text}.
\newblock In \emph{Proceedings of the Software Demonstrations of the 15th
  Conference of the {E}uropean Chapter of the Association for Computational
  Linguistics}, pages 49--52, Valencia, Spain. Association for Computational
  Linguistics.

\bibitem[{Prager(2019)}]{prager2019exploring}
Richard Heinz~Patrick Prager. 2019.
\newblock Exploring the use of role-playing games in education.
\newblock \emph{The MT Review}.

\bibitem[{Radford et~al.(2019)Radford, Wu, Child, Luan, Amodei, and
  Sutskever}]{radford2019language}
Alec Radford, Jeff Wu, Rewon Child, David Luan, Dario Amodei, and Ilya
  Sutskever. 2019.
\newblock Language models are unsupervised multitask learners.

\bibitem[{Raji et~al.(2021)Raji, Bender, Paullada, Denton, and
  Hanna}]{raji2021ai}
Inioluwa~Deborah Raji, Emily~M. Bender, Amandalynne Paullada, Emily Denton, and
  Alex Hanna. 2021.
\newblock \href {http://arxiv.org/abs/2111.15366} {Ai and the everything in the
  whole wide world benchmark}.

\bibitem[{Rajpurkar et~al.(2018)Rajpurkar, Jia, and Liang}]{rajpurkar2018know}
Pranav Rajpurkar, Robin Jia, and Percy Liang. 2018.
\newblock \href {http://arxiv.org/abs/1806.03822} {Know what you don't know:
  Unanswerable questions for squad}.

\bibitem[{Rombach et~al.(2021)Rombach, Blattmann, Lorenz, Esser, and
  Ommer}]{rombach2021highresolution}
Robin Rombach, Andreas Blattmann, Dominik Lorenz, Patrick Esser, and Björn
  Ommer. 2021.
\newblock \href {http://arxiv.org/abs/2112.10752} {High-resolution image
  synthesis with latent diffusion models}.

\bibitem[{Touvron et~al.(2023)Touvron, Lavril, Izacard, Martinet, Lachaux,
  Lacroix, Rozière, Goyal, Hambro, Azhar, Rodriguez, Joulin, Grave, and
  Lample}]{touvron2023llama}
Hugo Touvron, Thibaut Lavril, Gautier Izacard, Xavier Martinet, Marie-Anne
  Lachaux, Timothée Lacroix, Baptiste Rozière, Naman Goyal, Eric Hambro,
  Faisal Azhar, Aurelien Rodriguez, Armand Joulin, Edouard Grave, and Guillaume
  Lample. 2023.
\newblock \href {http://arxiv.org/abs/2302.13971} {Llama: Open and efficient
  foundation language models}.

\bibitem[{Trischler et~al.(2016)Trischler, Wang, Yuan, Harris, Sordoni,
  Bachman, and Suleman}]{trischler2016newsqa}
Adam Trischler, Tong Wang, Xingdi Yuan, Justin Harris, Alessandro Sordoni,
  Philip Bachman, and Kaheer Suleman. 2016.
\newblock {NewsQA: A machine comprehension dataset}.
\newblock \emph{arXiv preprint arXiv:1611.09830}.

\bibitem[{Van~den Berghe et~al.(2021)Van~den Berghe, Oudgenoeg-Paz, Verhagen,
  Brouwer, De~Haas, De~Wit, Willemsen, Vogt, Krahmer, and
  Leseman}]{van2021individual}
Rianne Van~den Berghe, Ora Oudgenoeg-Paz, Josje Verhagen, Susanne Brouwer,
  Mirjam De~Haas, Jan De~Wit, Bram Willemsen, Paul Vogt, Emiel Krahmer, and
  Paul Leseman. 2021.
\newblock Individual differences in children’s (language) learning skills
  moderate effects of robot-assisted second language learning.
\newblock \emph{Frontiers in Robotics and AI}, page 259.

\bibitem[{Volodina et~al.(2014)Volodina, Borin, and Pil{\'a}n}]{ws-2014-nlp}
Elena Volodina, Lars Borin, and Ildik{\'o} Pil{\'a}n, editors. 2014.
\newblock \href {https://aclanthology.org/W14-3500} {\emph{Proceedings of the
  third workshop on {NLP} for computer-assisted language learning}}. LiU
  Electronic Press, Uppsala, Sweden.

\bibitem[{Wei et~al.(2022)Wei, Bosma, Zhao, Guu, Yu, Lester, Du, Dai, and
  Le}]{wei2022finetuned}
Jason Wei, Maarten Bosma, Vincent~Y. Zhao, Kelvin Guu, Adams~Wei Yu, Brian
  Lester, Nan Du, Andrew~M. Dai, and Quoc~V. Le. 2022.
\newblock \href {http://arxiv.org/abs/2109.01652} {Finetuned language models
  are zero-shot learners}.

\bibitem[{Yang and Kyun(2022)}]{yang2022current}
Hongzhi Yang and Suna Kyun. 2022.
\newblock The current research trend of artificial intelligence in language
  learning: A systematic empirical literature review from an activity theory
  perspective.
\newblock \emph{Australasian Journal of Educational Technology},
  38(5):180--210.

\end{thebibliography}

\end{document}